\documentclass{article}
\usepackage{ijcai21}

\usepackage{times}
\usepackage{soul}
\usepackage{url}
\usepackage[hidelinks]{hyperref}
\usepackage[utf8]{inputenc}
\usepackage[small]{caption}
\usepackage{graphicx}
\usepackage{amsmath}
\usepackage{amsthm}
\usepackage{booktabs}
\usepackage{algorithm}
\usepackage{algorithmic}
\title{Tag, Copy or Predict: A Unified Weakly-Supervised Learning Framework for Visual Information Extraction using Sequences}

\author{
Jiapeng Wang$^1$
\and
Tianwei Wang$^1$\and
Guozhi Tang$^{1}$\and
Lianwen Jin$^{1,3}$\thanks{Corresponding author}\and
Weihong Ma$^1$\and\\
Kai Ding$^2$\And
Yichao Huang$^2$
\affiliations
$^1$School of Electronic and Information Engineering, South China University of Technology, China\\
$^2$IntSig Information Co.,  Ltd, Shanghai, China\\
$^3$Guangdong Artificial Intelligence and Digital Economy Laboratory (Pazhou Lab), Guangzhou, China
\emails
scutjpwang@foxmail.com,
wangtw@foxmail.com,
eetanggz@mail.scut.edu.cn,
eelwjin@scut.edu.cn,\\
eeweihong\_ma@mail.scut.edu.cn,
danny\_ding@intsig.net, charlie\_huang@intsig.net
}

\makeatletter
\newenvironment{breakablealgorithm}
  {
   \begin{center}
     \refstepcounter{algorithm}
     \hrule height.8pt depth0pt \kern2pt
     \renewcommand{\caption}[2][\relax]{
       {\raggedright\textbf{\fname@algorithm~\thealgorithm} ##2\par}%
       \ifx\relax##1\relax 
         \addcontentsline{loa}{algorithm}{\protect\numberline{\thealgorithm}##2}%
       \else 
         \addcontentsline{loa}{algorithm}{\protect\numberline{\thealgorithm}##1}%
       \fi
       \kern2pt\hrule\kern2pt
     }
  }{
     \kern2pt\hrule\relax
   \end{center}
  }
\makeatother

\begin{document}

\maketitle

\begin{abstract}
Visual information extraction (VIE) has attracted increasing attention in recent years. The existing methods usually first organized optical character recognition (OCR) results into plain texts and then utilized token-level entity annotations as supervision to train a sequence tagging model. However, it expends great annotation costs and may be exposed to label confusion, and the OCR errors will also significantly affect the final performance. In this paper, we propose a unified weakly-supervised learning framework called TCPN (Tag, Copy or Predict Network), which introduces 1) an efficient encoder to simultaneously model the semantic and layout information in 2D OCR results; 2) a weakly-supervised training strategy that utilizes only key information sequences as supervision; and 3) a flexible and switchable decoder which contains two inference modes: one ({\em Copy or Predict} Mode) is to output key information sequences of different categories by copying a token from the input or predicting one in each time step, and the other ({\em Tag} Mode) is to directly tag the input sequence in a single forward pass. Our method shows new state-of-the-art performance on several public benchmarks, which fully proves its effectiveness.

\end{abstract}

\begin{figure}[t!]
\centering
\begin{minipage}[]{0.98\linewidth}
\centering
\includegraphics[width=8.5cm,height=5cm]{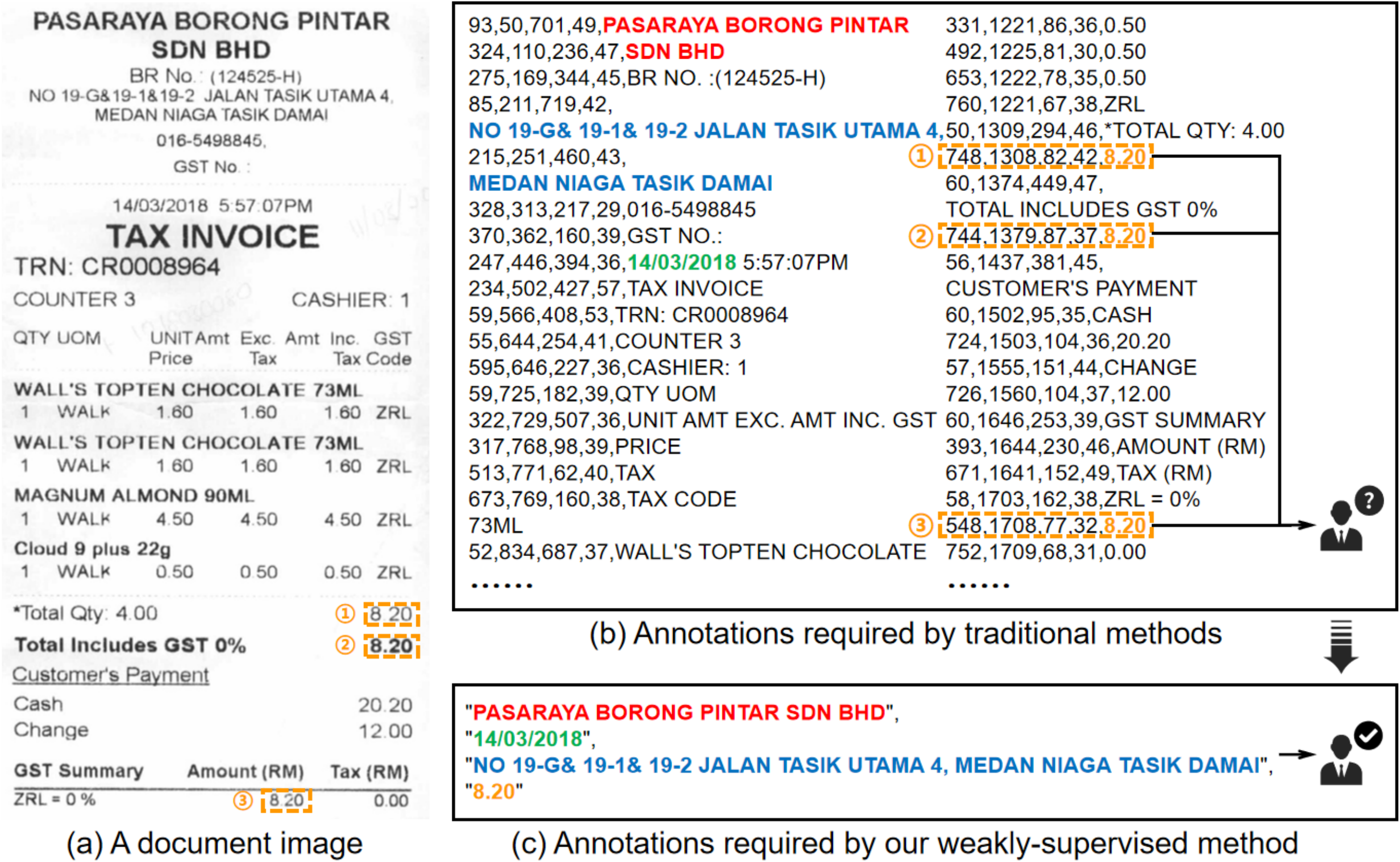}
\end{minipage}
\caption{Illustration of the annotations required by the traditional method and our weakly-supervised framework. Given (a) a document image, (b) traditional annotation scheme is to label the bounding box and string of each utterance, and further specific which category does each  token/box belongs to. In contrast, (c) our method only requires each key information sequence. Annotation ambiguity of traditional annotation scheme is shown in the orange dotted bounding boxes. Different colors denote different entity categories. }\label{fig:weaksuper}
\end{figure}

\section{Introduction}
With the fast development  of information interaction, document 
intelligent processing~\cite{intelligent} has attracted considerable attention.
As an important part of it, visual information extraction (VIE) technique has been integrated into many real-world applications.

The existing VIE methods usually first organized text blocks (text bounding boxes and strings, which were provided by the ground truth or parsed by an OCR system) into plain texts according to the reading order and utilized effective encoding structures such as ~\cite{Chargrid,gat_ali,layoutlm} to extract the most distinguishable representations for each input token from multi-sources. After this, a sequence tagging model like ~\cite{blstmcrf} was trained with token-level category supervision. 

However, the token-level category supervision expends great annotation costs and may be exposed to label ambiguity. Given a document image as shown in Figure \ref{fig:weaksuper}
(a), the most widely used annotation scheme is to label the bounding box and string of each utterance, and further point out which category does each token/box belongs to, as shown in Figure \ref{fig:weaksuper} (b). In this way, a heuristic label assignment procedure is needed to train the aforementioned tagging model,  of which the core idea is matching the detected boxes and recognized transcriptions with the given annotations and then assign label to each token/box of OCR results.
However, this procedure may encounter problems from mainly two aspects. First, wrong recognition results will bring troubles to the matching operation, especially for key information sequences. Second, the repeated contents will bring label ambiguities. As shown in Figure \ref{fig:weaksuper}(a) and (b), three values with same content can be regarded as the answer of the key {\em Total Amount}. In most cases, it is hard to establish a uniform annotation specification to determine which one should be regarded as ground truth.

To address the aforementioned limitations, in this paper, we propose an end-to-end weakly-supervised learning framework, which can supervise the decoding process directly  using the target key information sequences. The benefits it brings are mainly two-folds: first, it greatly saves the annotation costs, as shown in Figure \ref{fig:weaksuper}
(c), and shortens the training process by skipping the matching between OCR results and the ground truth; second, our method solves the label ambiguity problem by automatically learning the alignment between OCR results and ground truth, which can adaptively distinguish the most likely one in the repeated  contents. 
In addition, we also propose a flexible decoder, which is combined with our weakly-supervised training strategy and have two switchable modes -- {\em Copy or Predict} Mode (TCPN-CP) and {\em Tag} Mode (TCPN-T), to balance the effectiveness and efficiency. In TCPN-CP, our decoder can generate key information sequences
by {\em copying} a token from the input or {\em predicting} one in each time step, which can both retain novel contents in input and correct OCR errors. 
And in TCPN-T, our decoder can directly label each token's representations into a specific category in a single forward pass, which maintains the fast speed. It is notable that our decoder only needs to be trained once to work in different modes.

Besides, we propose an efficient encoder structure to simultaneously model the semantic and layout information in 2D OCR results. In our design, the semantic embeddings of different tokens in a document are re-organized into a vector matrix $I\in\mathcal{R}^{H\times W\times d}$ (here $H$, $W$ and $d$ are the height dimension,  the width dimension and the number of channels, respectively), which we called \textbf{TextLattice}, according to the center points of token-level bounding boxes. 
After this, we adopt a modified light-weight ResNet~\cite{resnet} combined with the U-Net~\cite{unet} architecture to generate the high-level representations for the subsequent decoding. It is notable that the most relevant work of our encoding method was CharGrid~\cite{Chargrid}, which first used CNN to integrate semantic clues in a layout. However, it initialized an empty vector of the same size as the original document image, and then repeatedly filled each token's embedding at every pixel within its  bounding box. This simple and direct approach may lead to the following limitations: 1) Larger tokens would be filled in more times, and it might result in the risk of category imbalance; 2) A pixel could completely represent a token, and repeated filling would waste extra cost; 3) The lack of concentration of information would make it difficult for the network to capture global clues.
By comparison, our method greatly saves computing resources while maintains the space information of document. For example, for a receipt image in SROIE~\cite{sroie} benchmark whose size is generally more than $1600 \times 800$, the side length of $I$ after our processing is less than 100. In this way, both holistic and local clues can be captured, the relative position relationship between utterances are retained, and distance can be perceived in a more intuitive way through the receptive field.

Experiments on the two public SROIE and EPHOIE~\cite{vies} benchmarks demonstrate that our method shows competitive and even new state-of-the-art performance.
Our main contributions can be summarized as follows:
\begin{itemize}
\item We propose an efficient 2D document representation called TextLattice, and the corresponding  light-weight encoder structure.
\item We propose a flexible decoder  which has two   inference modes - TCPN-CP for OCR error correction and TCPN-T for  fast inference speed.
\item We propose a weakly-supervised learning framework which guides  decoding process directly  using the key information sequences. This greatly saves the annotation costs and avoids label ambiguity.
\item Our method achieves competitive performances even compared with the setting of token-level full supervision, which totally proves its superior.
\end{itemize}

\section{Related Work}

\begin{figure*}[t]
\centering
\includegraphics[width=1.0\textwidth]{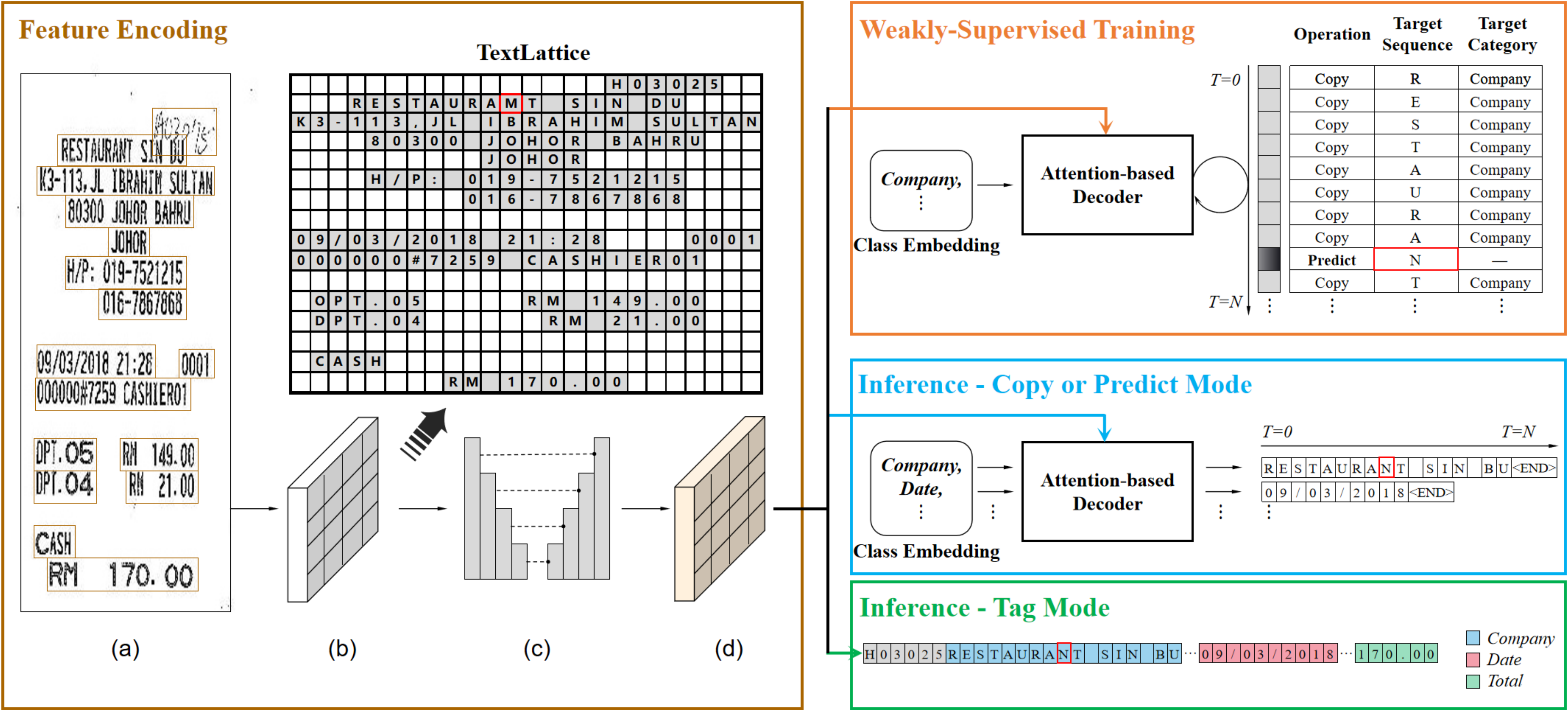}
\caption{The overall framework of TCPN. (a) The input OCR results (the recognition result  is ignored for visual clarity). (b) 
The corresponding TextLattice $I$.
(c) The light-weight modified ResNet encoder combined with the U-Net architecture. (d) The result of feature encoding, which has the same size as $I$. The class embeddings of multiple entity categories and the corresponding target sequences are given to the attention-based decoder during training. In inference, different modes can be switched to depends on application requirements. The red bounding box shows a wrongly recognized character `M'. In {\em Tag} Mode, it can be classified in a single forward pass; while in {\em  Copy or Predict} Mode, it can be corrected as `N'.
} 
\label{pipeline}
\end{figure*}

Early works of visual information extraction mainly utilized rule-based ~\cite{earlywork} or template-based~\cite{earlywork2} methods, which might tend to poor performance when the document layout was variable. With the development of deep learning, more advanced researches commonly extracted a feature sequence from the input plain text and used token-level supervision to train a sequence tagging model. \cite{blstmcrf} first used a bidirectional long short-term memory~\cite{lstm} (BiLSTM) network to model sequential information and a conditional random field (CRF) layer to decode the optimal classification path. Most of the follow-up works mainly focused on the feature encoding part: ~\cite{gat_ali}, GraphIE~\cite{Graphie} and  PICK~\cite{PICK}   tried to use graph neural networks (GNNs) to extract node embeddings for better representation. LayoutLM~\cite{layoutlm} embedded multi-source information into a common feature space, and utilized a BERT-like~\cite{bert} model for feature fusion. TRIE~\cite{TRIE} and VIES~\cite{vies} proposed the end-to-end VIE methods directly from image to key information, which introduced multi-head self-attention~\cite{attention} to integrate multimodal clues during encoding. 

As for the decoding part, EATEN~\cite{Eaten} first utilized sequence-level supervision to guide training. It generated feature maps directly from document image, and used several entity-aware attention-based decoders to iteratively parse the key information sequences. However, its efficiency could be significantly reduced as the number of entities increased, and it can only process simple and fixed layout since it had to overcome the difficulties of both OCR and VIE at the same time. When the given text blocks were accurate or directly the ground truth but the model still performed inference by step-by-step prediction, it might greatly slow down the speed and lead to the severe over-fitting problem of sequence generation due to the lack of corpus in VIE task.

\section{Method}
Here we provide the details of the proposed TCPN. First we describe the approach
of generating TextLattice,
and how to encode higher-level features.
Next we introduce details of our switchable decoder and weakly-supervised training strategy. 
Finally, we explain when and how to inference in different modes. 
Figure \ref{pipeline} gives an overview of our approach.

\subsection{Document Representation}
In this section, we introduce how to re-organize the OCR results into our 2D document representation - \textbf{TextLattice}. The whole process can be summarized as: 1) We first normalize $y$ coordinates of detected bounding boxes $B^u$, sort $B^u$ from top to bottom and left to right, and utilize heuristic rules to modify $y$ coordinates to
divide $B^u$ into multiple rows;
2) Then, $B^u$ is divided into token-level $B^t$ according to lengths of the recognized strings $S^u$; 3) Next, the $x$ coordinates of $B^t$ are also normalized and modified to avoid information loss caused by overlapping; 4) Finally, we initialize an all-zero matrix $I\in\mathcal{R}^{H\times W\times d}$ where $W$ and $H$ are determined by the ranges of $x$ and $y$ coordinates of $B^t$, and fill in $I$ according to the correspondence between token-level center points and $d$-dimensional token embeddings. The detailed procedure is shown  in Appendix.

\subsection{Feature Encoding}
After acquiring $I$, we adopt ResNet~\cite{resnet} as CNN encoder to capture more holistic  features. The U-Net~\cite{unet} structure is also combined to restore the down-sampled features to the same size as the input  $I$ and adaptively fuse both local and global clues extracted under diverse receptive fields. Vanilla ResNet adopts a $7 \times 7$ Conv2d as the {\tt conv1} layer to capture association between local pixels in an RGB image. However,  it may not be applicable in our scenario since the features of adjacent tokens also need to be separable, instead of high fusion of features in the first few layers. To this end, we replace {\tt conv1} with a $3 \times 3$ Conv2d and remove the original {\tt maxpool} layer. 
Thanks to the efficient design of TextLattice, both speed and superiority can be retained. 

In order to further preserve the location clues, inspired by CoordConv~\cite{coordconv}, two extra channels are concatenated to the incoming representation $I$, which contain horizontal and vertical relative location information in the layout of range from $-1$ to $1$. The whole procedure of feature encoding can be formulated as:
\begin{align}
    \hat{I} & = I + UNet(ResNet(I \oplus I_0)) \label{res1}\\
    F & = Indexing(\hat{I}, B^t)
\end{align}%
Here, $\oplus$ is the concatenation operator, $I_0\in\mathcal{R}^{H\times W\times 2}$ is the extra coordinate vector. Since the output of CNNs has the same size as $I$, we add them together as a residual connection. 
Finally, the features at the center points of token-level bounding boxes $B^t$ are retrieved to form $F\in\mathcal{R}^{N \times d}$, where $N$ is the number of tokens. We regard the rest pixels as useless and discard them directly for calculation efficiency.

\subsection{Weakly-Supervised Training}
As shown in Figure \ref{pipeline}, the VIE task can be regarded as a set-to-sequence problem after feature encoding, since $F$ is order-independent. We introduce the class embedding $C\in\mathcal{R}^{d}$ to control the category of information parsed by the decoder, which is taken from a pre-defined trainable look-up table. Given $C$, our attention-based decoder takes it into account at each time step and iteratively predicts target sequence.
Such novel design 
avoids class-specific decoders, alleviates the shortage of isolated class corpus, and decouples the serial correlation between different categories in the traditional sequence tagging model into parallel.

When generating sequences, we need the model to be able to switch between copying tokens from input or directly predicting ones. The copying operation make the model be able to reproduce accurate information and retain novel words, while the predicting operation introduces the ability of correcting OCR errors.  Inspired by ~\cite{cog}, which implemented a similar architecture for abstractive text summarization, our model  recurrently generates the hidden state $s_t$ by referring to the current input tokens $x_t$ and the context vector in the previous step $F_{t-1}^\ast$:
\begin{small}
\begin{align}
 e_i^t  & = W_etanh(W_1\textbf{C}+W_2F_i+W_3s_t+W_4\sum_{t{'}=1}^{t-1}\alpha_i^{t{'}}+b_1)\label{start1}\\
 {\alpha^t} & = Softmax({e^t}) \\
 F_t^\ast & = \sum_{i}{\alpha_i^tF_i} \\
 {s_t} & = RNN(F_{t-1}^\ast\oplus x_t,s_{t-1})\label{pred1}
\end{align}
\end{small}Here, $\alpha$ is the attention score where the historical accumulated values are also referenced during calculation. All $W$s and $b$s are learnable parameters.

Then, the probability distribution of tokens in a fixed dictionary $P_t^{pred}$ is calculated and a {\em copy score} $p_t^{copy}$ is generated as a soft switch to choose between different operations in each time step $t$:
\begin{small}
\begin{align}{
P_t^{pred} & =  Softmax(W_5(F_t^\ast\oplus s_t)+b_2) \\
p_t^{copy} & = \sigma(W_6F_t^\ast+W_7s_t+W_8x_t+b_3) \\
P_t(k^\ast) & = p_t^{copy}\sum_{i:k_i=k^\ast}\alpha_i^t+(1-p_t^{copy})P_t^{pred}(k^\ast) \label{pred7}\\
\mathcal{L}^{S}_t & = - log(P_t(k^\ast))
}\end{align}
\end{small}$P_t(k^\ast)$ is the probability score of token $k^\ast$ in time step $t$, where $k^\ast$ is the current target token. $\mathcal{L}^{S}_t$ is the sequence alignment loss function. In this way, our method  acquires the ability to produce out-of-vocabulary (OOV) tokens, and can adaptively perform optimal operations.

As of now, our method can be seen as a sequence generation model trained with sequence-level supervision. However, it is notable that since the class embedding $C$ of entity category $c$ is given, when the model decides to copy a token $k_i$ from the input at a step, $k_i$'s feature vector in $F$ should be also classified as $c$ by a linear classifier. 
More generally speaking, our method can first learn the alignment relationship, and then train a classifier using the matched tokens. This novel idea enables our approach the ability of supervising the sequence tagging model. We adopt a linear layer to model the entity probability distribution, which can be formulated as:
\begin{small}
\begin{align}{
P^{c}_{t,{i^*}} & = Softmax\left(W_cF_{k_{i^*}}+b_c\right), \label{c1}\\
where \qquad {i^*} & = \underset{i}{\rm argmax} (\alpha_i^t) \qquad and  \qquad p_t^{copy}>0.5. \label{cond} \\
\mathcal{L}^{C}_t & = - log(P^{c}_{t,{i^*}}(c)) \label{c3}
}\end{align}
\end{small}It is worth noting that, equation (\ref{c1}) - (\ref{c3}) do not train the tokens which do not belong to any key information sequences. The neglect of  negative samples may lead to severe defect that all input tokens will be classified as positive. Thus we construct an extra auxiliary loss function $\mathcal{L}_t^N$ for negative sample suppression:
\begin{small}
\begin{align}{
P^{c}_{t,i} &= Softmax\left(W_{c}F_{k_i}+b_c\right) \label{tag1}\\
P_t(c) &= \sum_{i}{P^{c}_{t,i}(c)}\\
\mathcal{L}_t^N &= max(0,\ P_t(c)\ -\ Length(S_c))
}\end{align}
\end{small}Here, $P_t(c)$ indicates the sum of the probabilities belong to entity category $c$ of all input tokens, and $Length(S_c)$ is the length of current target sequence $S_c$. The main purpose of $\mathcal{L}_t^N$ is to limit the number of input tokens classified as $c$ to be less than or equal to the actual number. This simple but effective design greatly improves the performance of the model in {\em Tag} Mode.

In summary, the final integrated loss function $\mathcal{L}_t$ is the weighted sum of multiple components mentioned above:
\begin{small}
\begin{align}{
\mathcal{L}_t &= \lambda_S \mathcal{L}^{S}_t  + \lambda_C \mathcal{L}^{C}_t + \lambda_N \mathcal{L}_t^N
}\end{align}
where $\lambda_S$, $\lambda_C$ and $\lambda_N$ are trade-off hyper-parameters.
\end{small}

\subsection{Inference}
In this section, we explain when and how to implement inference process in different modes. It is worth noting that, since class embeddings are sent into the decoder in the form of a batch, key information sequences of different categories of the same document can be generated under different modes according to the entity-specific semantic characteristics. 

In most real-world scenarios, OCR results cannot be flawless. In this regard, users can switch our decoder to {\em Copy or Predict} Mode as described in equation (\ref{start1}) - (\ref{pred7}) to
supplement missing or wrong tokens. This mode is more suitable for sequences of  categories with strong semantic relevance.

Thanks to the auto-alignment property of the proposed weakly-supervised training strategy, the decoder can also directly perform sequence tagging in a single forward pass in {\em Tag} Mode using equation (\ref{tag1}). It prefers to extremely few OCR errors or categories of weak semantic correlation between adjacent contents.

\section{Experiment}
\subsection{Implementation Details}
We adopt ResNet-18\cite{resnet} as backbone in feature encoding, and use BiLSTM in attention mechanism of decoder. The number of channels $d$ is set to 256, 
the hyper-parameters $\lambda_S$, $\lambda_C$ and $\lambda_N$ are all set to 1.0 in our experiments empirically. We set batch size as 4 and perform 450 training epochs. The  learning rate is initialized as 1.0  with ADADELTA~\cite{adadelta} optimization and decreased to a tenth  for two times in 300 and 400 epochs. 

\subsection{Ablation Study}
In this section, we evaluate the influences of  components of the proposed TCPN on the public EPHOIE benchmark.

\begin{table}
\centering
\begin{tabular}{lcc}
\toprule
\textbf{Encoding Architecture}  & \textbf{F1-Score} & \textbf{FPS} \\
\midrule
BiLSTM in \cite{blstmcrf}      & 96.16  & 103.84      \\
Chargrid\cite{Chargrid}      & 96.23  & 5.54      \\
GAT in \cite{gat_ali}    & 96.37  & 88.65      \\
BERT-like\cite{cbert}   & 97.19  & 62.23      \\
\textbf{TextLattice(Ours)} & \textbf{98.06}  & \textbf{112.11}      \\
\bottomrule
\end{tabular}
\caption{Performance and speed comparison on EPHOIE dataset between different encoding architectures. FPS is tested on a GeForce GTX 1080 Ti.}
\label{tab:encoding}
\end{table}

\begin{table}
\centering
\begin{tabular}{lc}
\toprule
\textbf{Encoding Architecture}  & \textbf{F1-Score}  \\
\midrule
\textbf{TextLattice(Ours)}\dag      & \textbf{98.06}   \\
\dag - CoordConv\cite{coordconv}    & 96.83    \\
\dag - UNet\cite{unet}   & 92.37      \\
\dag - Residual Connection & 97.70    \\
\bottomrule
\end{tabular}
\caption{Effects of different components in the proposed encoding architecture on EPHOIE dataset.}
\label{tab:enabs}
\end{table}

\subsubsection{Comparison between Different Encoding Structures}
We organize OCR ground truth into plain texts and use different encoding structures to generate representations. Then a sequence tagging model is trained utilizing the official token-level category annotations. We mainly adopt the following models for comparison:
 \textbf{BiLSTM}: A bidirectional LSTM  adopted in \cite{blstmcrf};
 \textbf{GAT}: A graph attention network (GAT) adopted in \cite{gat_ali};
 \textbf{BERT-like}: A BERT-like model similar to LayoutLM\cite{layoutlm}. Since vanilla LayoutLM is trained on English corpus, the pretrained weights for Chinese in \cite{cbert} are loaded for fair comparison;
 \textbf{Chargrid}: The document modeling method introduced by Chargrid\cite{Chargrid}.

The comparison results are given in Table \ref{tab:encoding}.  \textbf{BiLSTM} perceives sequential clues well, but it cannot effectively model location space in  1D form; \textbf{GAT} can adaptively fuse useful features using attention mechanism. However, the ability of capturing  location clues highly depends on the way of  feature embedding; \textbf{BERT-like} can perform forward calculation in parallel, and since the pretrained weights are loaded, it achieves satisfactory performance; \textbf{Chargrid} establishes input matrix using a more direct way, which means that both robustness and efficiency cannot be guaranteed. It is notable that \textbf{TextLattice(Ours)} achieves superior performance and  maintains the fastest speed, which fully proves its efficiency. Our method has a more direct and sensitive perception of location clues than position embedding in \textbf{GAT} or \textbf{BERT-like}, and ensures a higher degree of information concentration than \textbf{Chargrid}. The fully parallel scheme also greatly contributes to the leading speed.



\begin{table}[t]
\centering
\begin{tabular}{lc}
\toprule
\textbf{Method}  & \textbf{F1-Score} \\
\midrule
\midrule
Token-level Fully-Supervised\\
\midrule
\cite{blstmcrf}       & 89.10       \\
\cite{gat_ali}    & 92.55    \\
GraphIE\cite{Graphie}   & 90.26       \\
TRIE\cite{TRIE}   & 93.21       \\
VIES\cite{vies}   & 95.23       \\
\textbf{TextLattice(Ours)} & \textbf{98.06}        \\
\midrule
Sequence-level Weakly-Supervised\\
\midrule
\textbf{TCPN-T(Ours)} & \textbf{97.59}\\
\bottomrule
\end{tabular}
\\(a)
\bigskip 

\begin{tabular}{lc}
\toprule
\textbf{Method}  & \textbf{F1-Score} \\
\midrule
\midrule
Token-level Fully-Supervised\\
\midrule
\cite{blstmcrf} & 90.85    \\
LayoutLM\cite{layoutlm}   & 95.24    \\
\cite{gat_ali}      &    95.10      \\
PICK \cite{PICK}       & 96.12    \\
TRIE \cite{TRIE}    & 96.18  \\ 
VIES\cite{vies}     &    96.12  \\
\textbf{TextLattice(Ours)} & \textbf{96.54}        \\
\midrule
Sequence-level Weakly-Supervised\\
\midrule
\textbf{TCPN-T(Ours)} & \textbf{95.46}\\
\bottomrule
\end{tabular}\\(b)
\caption{Performance comparison on (a) EPHOIE and (b) SROIE under ground truth setting. `Token-level Fully-Supervised' indicates the token-level category annotation, and `Sequence-level Weakly-Supervised' means the key information sequence annotation.}
\label{tab:ephoiegt}
\end{table}

\subsubsection{Effects of Components in TextLattice}
We also conduct experiments to verify the effectiveness of different components in  our encoding structure, such as CoordConv, the U-Net structure and the residual connection in equation (\ref{res1}). It can be seen in Table \ref{tab:enabs} that each design has a significant contribution to the final performance. Although CNN can capture the relative position relationship, \textbf{CoordConv} can further provide the global position clues relative to the whole layout, which brings higher discernibility; we also try to use \textbf{ResNet only} where all {\tt stride} and the U-Net structure are removed to perform feature encoding. However, the performance decreases obviously, which  indicates  the importance of semantic feature fusion under different receptive fields; \textbf{Residual Connection} gives model the  chance to directly receive token-level semantic embedding, which further improves the performance.

\subsection{Comparison with the State-of-the-Arts}
We compare our method with several state-of-the-arts on the SROIE and EPHOIE benchmarks. The following \textbf{Ground Truth Setting} indicates that the  OCR ground truth is adopted, while \textbf{End-to-End Setting} indicates the OCR engine result. 

\begin{table}[t]
\centering
\begin{tabular}{lc}
\toprule
\textbf{Method}  & \textbf{F1-Score} \\
\midrule
\midrule
Rule-based Matching\\
\midrule
\cite{blstmcrf}       & 71.95       \\
\cite{gat_ali}    & 75.07    \\
TRIE\cite{TRIE}   & 80.31       \\
VIES\cite{vies}   & 83.81       \\
\midrule
Automatic Alignment\\
\midrule
\textbf{TCPN-T(Ours, 112.11FPS)} & \textbf{86.19}        \\
TCPN-CP(Ours, 5.57FPS) & 84.67 \\
\bottomrule
\end{tabular}
\\(a)
\bigskip 

\begin{tabular}{lc}
\toprule
\textbf{Method}  & \textbf{F1-Score} \\
\midrule
\midrule
Rule-based Matching\\
\midrule
NER \cite{ner}         & 69.09    \\
Chargrid \cite{Chargrid}       & 78.24    \\  \cite{blstmcrf}      &    78.60      \\
\cite{gat_ali}         &       80.76   \\
TRIE \cite{TRIE}          & 82.06    \\
VIES \cite{vies}  & 91.07 \\
\midrule
Automatic Alignment\\
\midrule
TCPN-T(Ours, 88.16FPS) & 91.21        \\
\textbf{TCPN-CP(Ours, 5.20FPS)} & \textbf{91.93} \\
\bottomrule
\end{tabular}\\(b)
\caption{Performance comparison on (a) EPHOIE and (b) SROIE under end-to-end setting. `Rule-based Matching' indicates acquiring token-level label through traditional rule-based matching, and `Automatic Alignment' means automatically learning the alignment using the key information sequences.}
\label{tab:sroieocr}
\end{table}

\subsubsection{Results under Ground Truth Setting}
As shown in Table \ref{tab:ephoiegt}, our method exhibits superior performance on both SROIE and EPHOIE in the case of token-level full-supervision, which totally proves the effectiveness of our feature encoding method. Furthermore, the results under sequence-level weakly-supervised setting achieve competitive performance. This fully confirms the superiority of our learning strategy, which can model the correspondence between the input tokens and the output sequences.

Compared with SROIE, EPHOIE usually has less content and more token types, which reduces the difficulty of learning alignment.
Relatively, since a receipt in SROIE often contains abundant tokens and the same token may appear repeatedly, which may lead to the alignment confusion,
the gap between full- and weak-supervision is further widened.

\subsubsection{Results under End-to-End Setting}
We adopt BDN\cite{bdn} as text detector and CRNN\cite{crnn} as text recognizer, and train them using the official annotations to get OCR results. It is worth noting that since there may inevitably exist errors in OCR results, all of our experiments under end-to-end setting are trained using our weakly-supervised manner, which avoids the matching process between OCR results and ground truth.
The performances are shown in Table \ref{tab:sroieocr}. Our method shows new state-of-the-art performance in every mode. 

It can be inferred that an important basis for choosing TCPN-CP or TCPN-T mode is the richness of semantics and corresponding corpus. On SROIE, TCPN-CP obviously outperforms TCPN-T, which mainly benefits the ability to error correction; however, on EPHOIE, although both modes outperform counterparts, TCPN-T beats TCPN-CP by a large margin, where the main reason should be the diversity of Chinese tokens and the resulting lack of corpus. 

\begin{table}
\centering
\begin{tabular}{lc}
\toprule
\textbf{Method}  & \textbf{F1-Score}  \\
\midrule
\midrule
Rule-based Matching\\
\midrule
\cite{blstmcrf}       & 78.79       \\
\cite{gat_ali}    & 80.92    \\
\textbf{TCPN-T(Ours)}   & \textbf{82.15}       \\
\midrule
Automatic Alignment\\
\midrule
TCPN-T(Ours)   & 84.37       \\
\textbf{TCPN-CP(Ours)}   & \textbf{89.08}    \\
\bottomrule
\end{tabular}
\caption{Performance comparison on Business License Dataset under end-to-end setting.}
\label{tab:bus}
\end{table}

\subsubsection{Results on A Business License Dataset}
In order to further explore the effectiveness of our framework in real-world  applications, we collect an in-house dataset of business license. It contains 2331 photos taken by mobile phone or camera with real user needs, and most of images are inclined, distorted or the brightness changes dramatically. We randomly select  1863 images for training and 468 images for testing,  and there are 13 types of entities to be extracted. Furthermore, the OCR results are generated by our off-the-shelf engines, which definitely contains OCR errors due to the poor image quality.

The detailed results are shown in Table \ref{tab:bus}. Our end-to-end weakly-supervised learning framework outperforms traditional rule-based matching method by a large margin, which can also greatly reduce the annotation cost. Compared with TCPN-T, the implicit semantic relevance learned by TCPN-CP can further boost the final performance by correcting OCR errors. Some qualitative results are shown in Appendix.

\section{Conclusion}
In this paper, we propose a unified weakly-supervised learning framework called TCPN for visual information extraction, which introduces an efficient encoder, a novel training strategy and a switchable decoder. Our method shows significant gain on EPHOIE dataset and competitive performance on SROIE dataset, which fully verifies its effectiveness.

Visual information extraction task is in the cross domain of  natural language processing and computer vision, and our approach aims to alleviate the over-reliance on complete annotations and the negative effects caused by OCR errors. For future research, we will explore the potential of our framework through large-scale unsupervised data. In this way, the generalization of  encoder, the alignment capability of  decoder and the performance of our TCPN-CP can be further improved.

\section*{Acknowledgments}
This research is supported in part by NSFC (Grant No.: 61936003, 61771199),  GD-NSF (no.2017A030312006).

\clearpage
\section*{Appendix}
\appendix
\section{TextLattice Generation}
Algorithm 1 shows the detailed procedure of our TextLattice Generation Algorithm introduced in Section 3.1.

\section{Qualitative Results}
\begin{figure}[t!]
\begin{minipage}[]{0.95\linewidth}
\centering
\includegraphics[width=7.2cm,height=20.5cm]{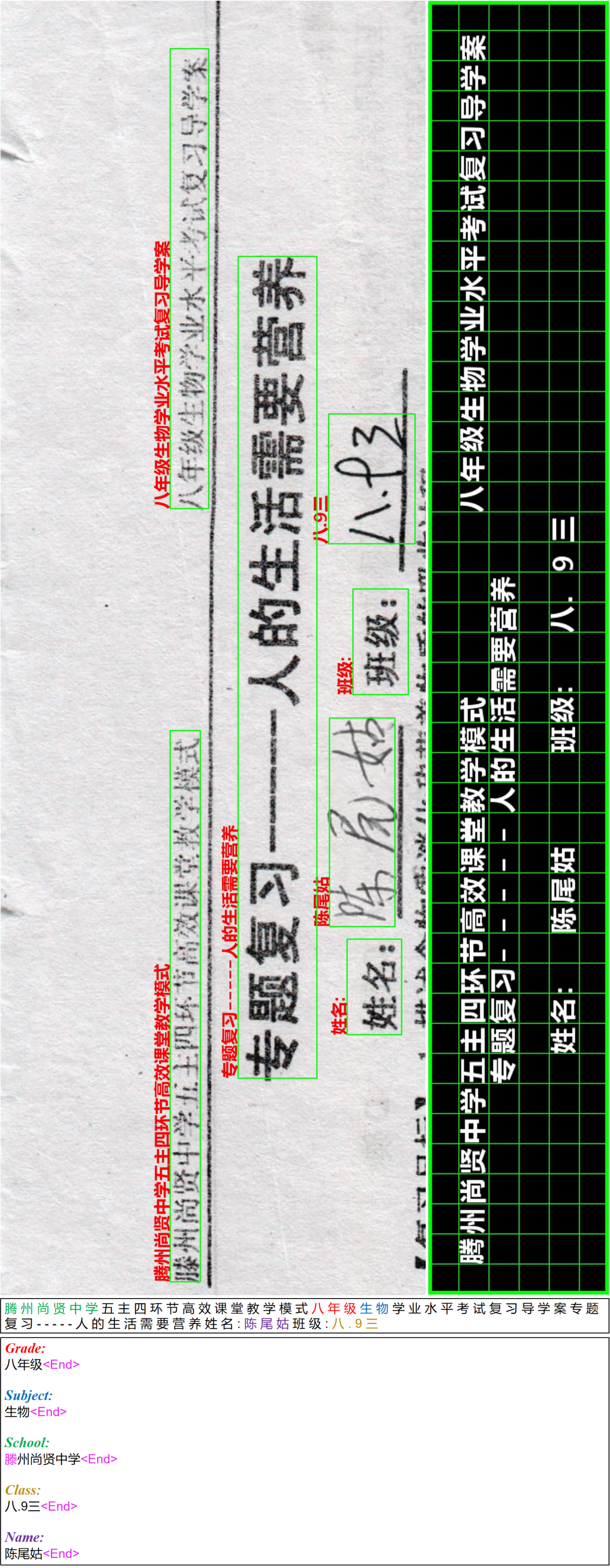}
\end{minipage}
\caption{Qualitative results of TCPN on EPHOIE dataset. From top to bottom and left to right are the OCR results, the TextLattice $I$, and the final predictions of TCPN-T and TCPN-CP. In TextLattice $I$, each green box represents a pixel. Pink tokens in TCPN-CP are generated by {\em predicting}, while the rest is produced by {\em copying}. Other different colors denote different entities.}\label{fig:res1}
\end{figure}

\begin{figure}[t!]
\begin{minipage}[]{0.95\linewidth}
\centering
\includegraphics[width=7cm,height=20.5cm]{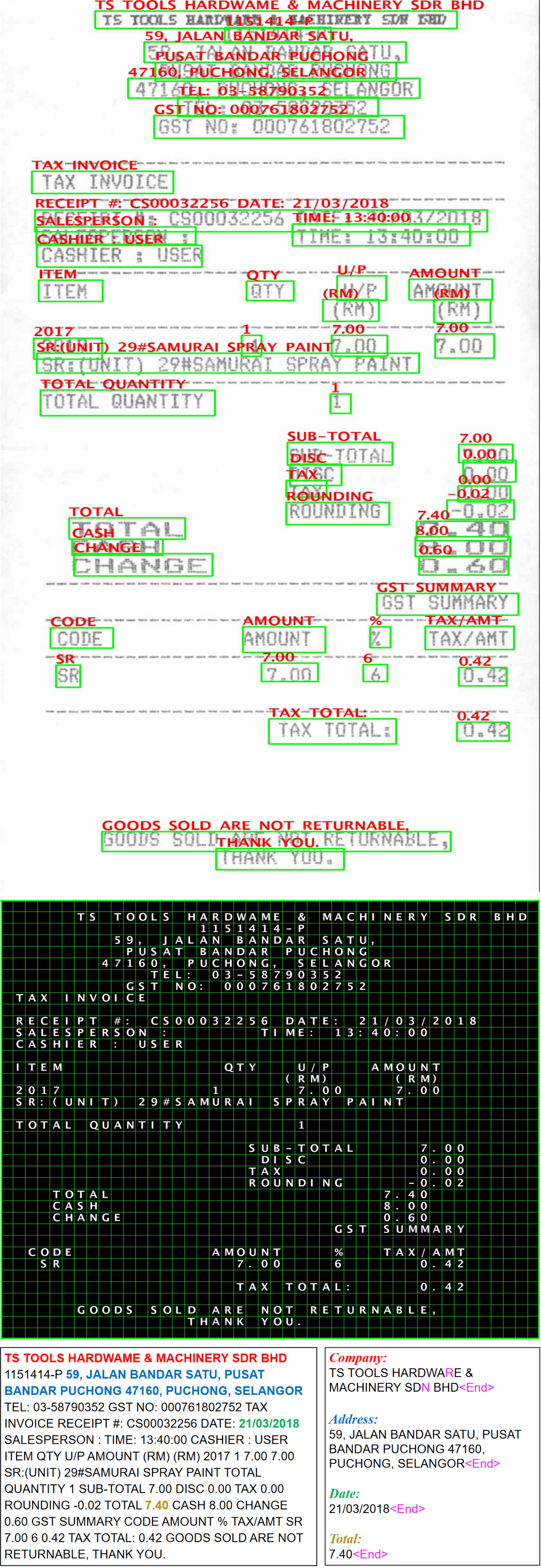}
\end{minipage}
\caption{Qualitative results of TCPN on SROIE dataset.}\label{fig:res2}
\end{figure}

\begin{figure}[t!]
\begin{minipage}[]{0.95\linewidth}
\centering
\includegraphics[width=7cm,height=20cm]{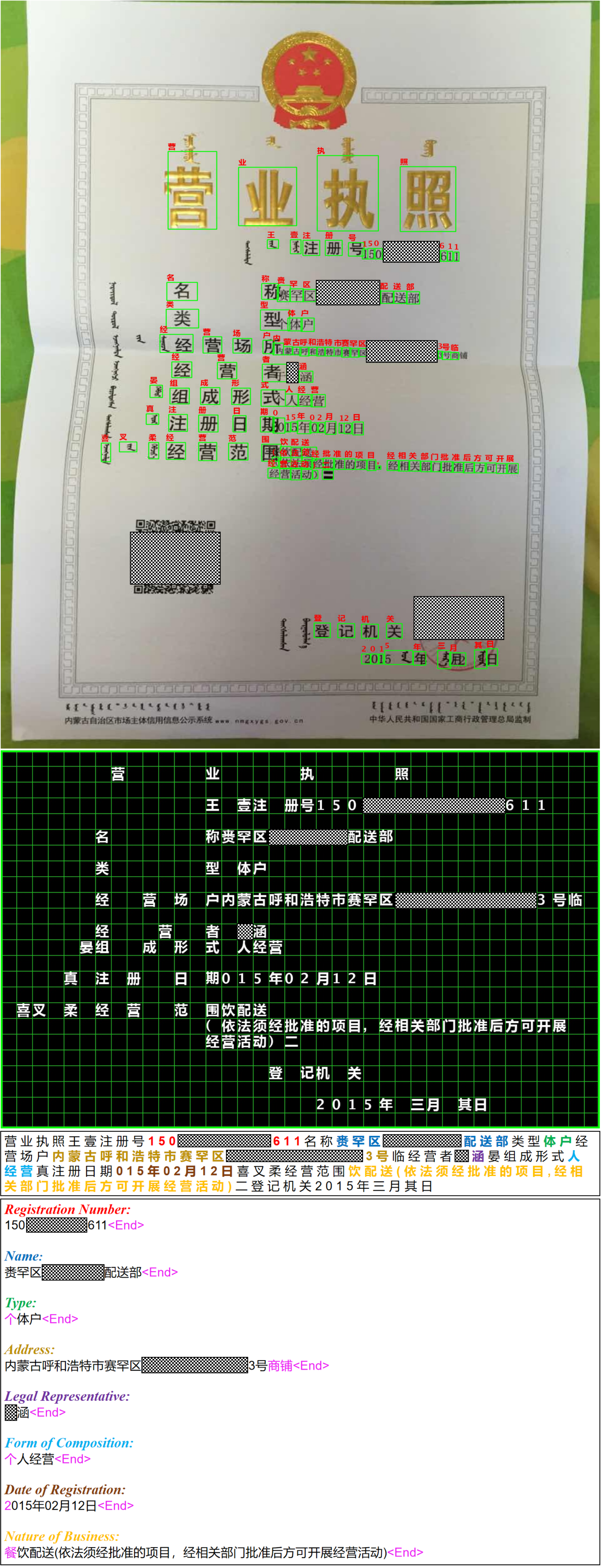}
\end{minipage}
\caption{Qualitative results of TCPN on the in-house Business License dataset. For the purpose of privacy protection, part of the correctly parsed information is {\em masked}.}\label{fig:res3}
\end{figure}

Some qualitative results are shown in Figure \ref{fig:res1}, \ref{fig:res2} and  \ref{fig:res3}.

\quad

\begin{breakablealgorithm}
\caption{TextLattice Generation Algorithm}
\raggedright \textbf{Input}: Utterance-level detected bounding boxes $B^u$ and corresponding recognized strings $S^u$ \\
\textbf{Parameter}: Normalization threshold $R^t$ and ratio $R^r$. \\
\textbf{Output}: TextLattice $I$ \\
\begin{algorithmic}[1] 
\STATE Calculate $R^h$ = Average height of $B^u$.
\STATE Normalize $y$ coordinates of $B^u$ using $R^h$.
\STATE Sort $B^u$ from top to bottom, and then from left to right.
\STATE Let $Y$ = unique values of $y$ coordinates of center points of $B^u$ sorted from small to large, a list $\widetilde{Y} = {[}Y[1],{]}$.
\FOR{$i$ from $2$ to Count($Y$)}
\IF {($Y[i] - Y[i-1]) \le R^t$}
\STATE $dY = 1$.
\ELSE
\STATE $dY = max(1, (Y[i] - Y[i-1]) / R^r)$.
\ENDIF \\
\STATE Append $\widetilde{Y}[-1] + dY$ into $\widetilde{Y}$. \\
\ENDFOR\\
\STATE Modify $y$ coordinates of center points of $B^u$ corresponding to $Y \rightarrow \widetilde{Y}$. \\ 
\STATE Split utterance-level boxes $B^u$ into token-level boxes $B^t$ according to lengths of $S^u$.\\
\STATE Calculate $R^w$ = Average width of $B^t$.\\
\STATE Normalize $x$ coordinates of $B^t$ using $R^w$.\\
\FOR{$\hat{y}$ in $Y$}
\STATE Let $\hat{X}$ = $x$ coordinates of boxes $B^t{'}$ whose $y$ coordinates of center points equal to $\hat{y}$.\\
\STATE Sort $\hat{X}$ from small to large, let $\widetilde{X} = {[}\hat{X}[1],{]}$\\
\FOR{$i$ from $2$ to Count($\hat{X}$)}
\IF {($\hat{X}[i] - \hat{X}[i-1]) \le R^t$}
\STATE $dX = 1$.
\ELSE
\STATE $dX = max(1, (\hat{X}[i] - \hat{X}[i-1]) / R^r)$.
\ENDIF \\
\STATE Append $\widetilde{X}[-1] + dX$ into $\widetilde{X}$. \\
\ENDFOR\\
\STATE Modify $x$ coordinates of center points of $B^t{'}$ corresponding to $\hat{X} \rightarrow \widetilde{X}$. \\ 
\ENDFOR\\
\STATE Combine $S^u$ in the same order as $B^u$ to get $S^t$.\\
\STATE Let $x_{min}, x_{max}, y_{min}, y_{max}$ = minimal and max values of $x$ and $y$ coordinates of center points of $B^t$, initialize an all-zero matrix $I \in\mathcal {R}^{(y_{max} - y_{min})\times (x_{max} - x_{min})\times d}$. Subtract $x_{min}, y_{min}$ from the coordinates of $B^t$.\\
\STATE Fill in $I$ according to the correspondence between $B^t$ and $d$-dimensional token embeddings of $S^t$. \\
\STATE \textbf{return} $I$.\\
\end{algorithmic}
\end{breakablealgorithm}

\clearpage


\bibliographystyle{named}
\bibliography{ijcai21}

\end{document}